\documentclass{sigkddExp}

\def\S{\mathbf{S}}

\def\y{\mathbf{y}}

\def\v{\mathbf{v}}

\def\x{\mathbf{x}}

\def\p{\mathbf{p}}

\def\R{\mathbf{R}}

\def\w{\mathbf{w}}
\def\t{\mathbf{t}}

\def\y{\mathbf{y}}

\def\w{\mathbf{w}}

\def\x{\mathbf{x}}
\def\P{\mathbf{P}}
\def\R{\mathbf{R}}

\def\Re{\mathbb{R}}
\def\0{\mathbf{0}}

\def\d{\mathbf{d}}

\begin{document}

\title{A Probabilistic Disease Progression Model for Predicting Future Clinical Outcome}

\numberofauthors{2}

\author{
%
\alignauthor Yingying Zhu \\
       \affaddr{Electrical and Computer Eng, Cornell}\\
       \affaddr{Ithaca, NY, USA}\\
       \email{yz2377@cornell.edu}
\alignauthor Mert R. Sabuncu\\
       \affaddr{Electrical and Computer Eng, Cornell}\\
       \affaddr{Ithaca, NY, USA}\\
       \email{msabuncu@cornell.edu}
}

\maketitle
\begin{abstract}

In this work, we consider the problem of predicting the course of a progressive disease, such as cancer or Alzheimer's.
Progressive diseases often start with mild symptoms that might precede a diagnosis, and each patient follows their own trajectory.
Patient trajectories exhibit wild variability, which can be associated with many factors such as genotype, age, or sex.
An additional layer of complexity is that, in real life, the amount and type of data available for each patient can differ significantly.
For example, for one patient we might have no prior history, whereas for another patient we might have detailed clinical assessments obtained at multiple prior time-points.
This paper presents a probabilistic model that can handle multiple modalities (including images and clinical assessments) and variable patient histories with irregular timings and missing entries, to predict clinical scores at future time-points.
We use a sigmoidal function to model latent disease progression, which gives rise to clinical observations in our generative model.
We implemented an approximate Bayesian inference strategy on the proposed model to estimate the parameters on data from a large population of subjects.
Furthermore, the Bayesian framework enables the model to automatically fine-tune its predictions based on historical observations that might be available on the test subject.
We applied our method to a  longitudinal Alzheimer's disease dataset with more than 3000 subjects \cite{adni} and present a detailed empirical analysis of prediction performance under different scenarios, with comparisons against several benchmarks.
We also demonstrate how the proposed model can be interrogated to glean insights about temporal dynamics in Alzheimer's disease.

\end{abstract}

\section{Introduction}


Many progressive disorders, such as Alzheimer`s disease (AD) \cite{AD2010} or cancer~\cite{cancer2013}, begin with mild symptoms that often precede diagnosis, and follow a patient-specific clinical trajectory that can be influenced by genetic and/or other factors. Therapeutic interventions, if available, are usually more effective in the earliest stages of a progressive disease~\cite{ADearly}. Therefore, tracking and predicting disease progression, particularly during the mild stages, is one of the primary objectives of personalized medicine.

In this paper, we are motivated by the real-world clinical setting where each individual is at risk and thus monitored for a specific progressive disease, such as AD. Furthermore, we assume that each individual might pay zero, one, or more visits to the clinic. In each clinical visit, various biomarkers or assessments (correlated with the disease and/or its progression) are obtained. Example biomarker modalities include brain MRI scans, PET scans, blood tests, and cognitive test scores. The number and timing of the visits, and the exact types of data collected at each visit can be planned to be standardized, but often vary wildly between patients in practice \cite{bernal2013}. An ideal clinical prediction tool should be able to  deal with this heterogeneity and compute accurate forecasts for arbitrary time horizons.

We present a probabilistic disease progression model that elegantly handles the aforementioned challenges of longitudinal clinical settings: data missingness, variable timing and number of visits, and multi-modal data (i.e., data of different types). The backbone of our model is a latent sigmoidal curve that captures the dynamics of the unobserved pathology, which is reflected in time-varying clinical assessments. Sigmoid curves are conceptually useful abstractions that fit well a wide range of dynamic physical and biological phenomena, including disease progression~\cite{Jack2010,stock2003,sabuncu2011}, which exhibit a floor and ceiling effect. In our framework, the sigmoid allows us to model the temporal correlation in longitudinal measurements and capture the dependence between the different tests and assessments, which are assumed to be generated conditionally independently from the latent state.
We implemented an approximate Bayesian inference strategy on the proposed model and applied it to a large-scale longitudinal Alzheimer's disease dataset \cite{adni}, a devastating, terminal neurodegenerative disease that affects over 10\% of the population older than 65.

In our experiments, we considered three target variables, which are widely used cognitive and clinical assessments associated with AD: the Mini Mental State Examination (MMSE)~\cite{mmse}, the Alzheimer's Disease Assessment Scale Cognitive Subscale (ADAS-COG)~\cite{cano2010adas}, and the Clinical Dementia Rating Sum of Boxes (CDR-SB)~\cite{o2008staging}.
 We trained and evaluated the proposed model on a longitudinal dataset with more than 3,000 subjects that included healthy controls (cognitively normal elderly individuals), subjects with mild cognitive impairment (MCI, a clinical stage that indicates high risk for dementia), and patients with AD. These clinical classifications are done in part based on the three target biomarkers in our model: MMSE, ADAS-Cog, and CDR-SB. Over the 5+ year follow-up period of the study, many of the subjects transitioned between clinical categories, e.g., from healthy to MCI or from MCI to AD.
 In fact, predicting these transitions is a significant focus of prior literature.
 Yet, the emphasis in prior work is in what might be called static prediction: the target variable is pre-fixed, e.g.,~\cite{daoqiangzhang2011,moradi2015machine}.
 For example, a common objective is to discriminate MCI subjects who progress to AD within some time window, e.g., 2 years.
 Similar problem setups are popular in cancer research, where a question of interest might be the 5 year survival.
 This is a binary classification problem.
 Our objective, on the the other hand, is to model and predict the entire clinical trajectory.
 Furthermore, we want to achieve this with a model that enables knowledge discovery about the progression of the disease.

We provide a detailed analysis of prediction accuracy achieved with the proposed model and alternative benchmark methods under different scenarios that involve varying the past available visits and future time windows.
In all our comparisons, the proposed model achieves significantly and substantially better accuracy for all target biomarkers.
Furthermore, due to the probabilistic and generative nature of our model, we are able to make certain mechanistic queries to gain further insights about the underlying dynamics.
This perspective offers us some new insights.
For instance, we provide a quantification of the impact of AD risk factors (such as APOE genotype) on disease progression.

The rest of this paper is organized as follows. Section 2 presents the proposed model and inference method. Section 3 describes the data and experimental set-up. Then, we present empirical results in Section 4. Finally, we conclude in Section 5.

\section{Proposed Method}
\begin{figure}[h]
\includegraphics[width=8cm]{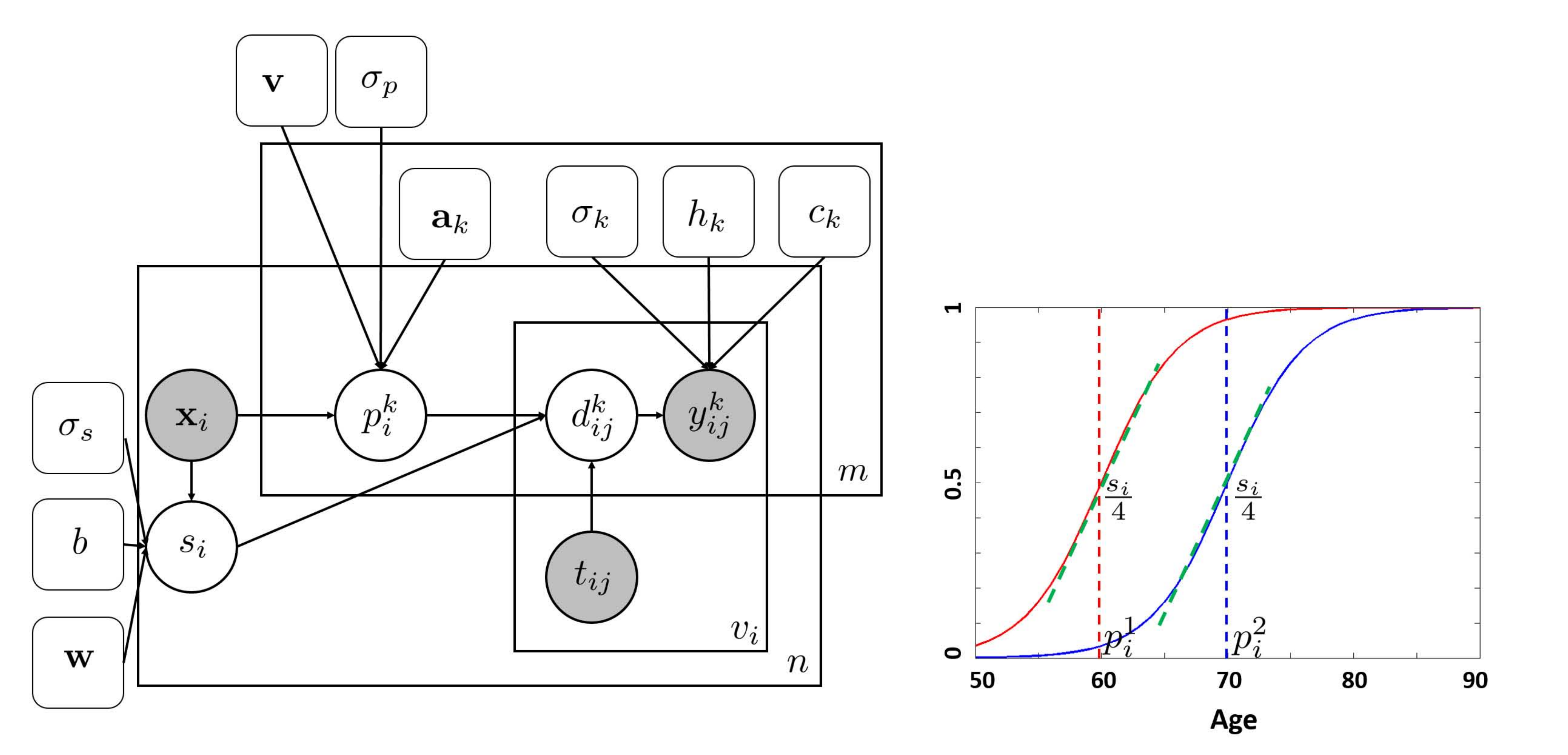}
\caption{
Left: Graphical model (Bayesian network) that depicts the assumed statistical dependency structure between random variables (circles) and model parameters (rectangles). We use standard conventions: shaded random variables are assumed to be observed during training, and plates indicate replication with the number of copies listed at the lower left corner. See text for information on variables.
Right: Illustration of sigmoids (with two different inflection points) that are assumed to model latent progression curves of two target variables.}
\label{Fig:dependency}
\end{figure}
\subsection{Model}

Let us first describe our notation and present our model. 
Assume we are given $n$ subjects.
 $\x_i \in \Re^{d\times 1}$ denotes subject $i$'s $d$-dimensional attribute vector.
 In our experiments, this vector contains APOE genotype (encoded as number of E4 alleles, which can be $0, 1$ or $2$)~\cite{corder1993gene}, education (in years)~\cite{katzman1993education}, sex (0 for female and 1 for male)~\cite{fratiglioni1991prevalence} and two well-established neuroanatomical biomarkers of AD computed from a baseline MRI scan (namely total hippocampal~\cite{jack1999prediction} and ventricular volume~\cite{nestor2008ventricular} normalized by brain size).
 The MRI biomarkers capture so-called ``brain reserve''~\cite{stern2012cognitive}.
Let $\y_i^k \in \Re^{v_i \times 1}$  represent the values of the the $k$'th dynamic (i.e., time-varying) target variable
at $v_i$ different clinical visits.
$\t_i = [t_{i1},\cdots, t_{iv_i}]\in \Re^{v_i\times 1}$ denotes a vector of the age of subject $i$ at these visits.
Note that the number and timing of the visits can vary across subjects.
In general, we will assume $k \in \{1, \cdots, m \}$.
In our experiments, we consider 3 target variables: MMSE, ADAS-COG or CDRSB and thus $m=3$.
We use $\d_{i}^k = [d_{i1}^k,\cdots, d_{iv_i}^k]$ to denote subject $i$'s latent trajectory values associated with the $k$'th target variable.
We assume each $d_{ij}^k \in [0, 1]$, with lower values corresponding to milder stages.
As we describe below, the target variable, which is a clinical assessment, will be assumed to be a noisy observation of this latent variable.
We model the latent trajectory of $\d_{i}^k$ as a sigmoid function of time (i.e., age), parameterized by a target- and subject-specific inflection point $p_i^k \in \Re$ and a subject-specific slope parameter $s_i \in \Re$.
Note that we assume that the slopes of the latent sigmoids associated with each target are coupled for each subject, yet the inflection points differ, which correspond to an average lag between the dynamics of target variables.
This is consistent with the hypothesized biomarker trajectories of AD~\cite{Jack2010}. However, it would be easy to relax this assumption by allowing each target variable to have its own slope.

We assume the inflection points $\{p_i^k \}$ and slopes $\{ s_i \}$  are random variables drawn from Gaussian priors with means equal to linear functions of subject-specific attributes $\x_i$:
\begin{eqnarray}
&& p_i^k \sim \mathcal{N}(\v^T\x_i+a_k, \sigma_p^2) \label{eq:p}\\
&& s_i \sim  \mathcal{N}({\w}^T{\x}_i+b, \sigma_s^2) \label{eq:s},
\end{eqnarray}
where $a_k \in \Re$ is associated with the $k$'th target (accounting for different time lags between target dynamics), while $\v, \w\in \Re^{d\times 1}$, and  $b, \sigma_p, \sigma_s \in \Re$ are general parameters.
Here and henceforth $\mathcal{N}(\mu, \sigma^2)$ denotes a Gaussian with mean $\mu$ and variance $\sigma^2$.

Given $s_i$ and  $p_i^k$, the latent value $d_{ij}^k$ associated with the $k$'th target is computed by evaluating the sigmoid at $t_{ij}$:
\begin{eqnarray}
&& d_{ij}^k= \frac{1}{1+\exp(-(t_{ij} - p_i^k ) s_i)}. \nonumber
\end{eqnarray}
The inflection point $p_i^k$ marks the age at which the rate of change achieves its maximum, which is equal to $s_i/4$.

Finally, we assume that the target variable value $y_{ij}^k$ is a linear function of the latent state $d_{ij}^k$ corrupted by additive zero-mean independent Gaussian noise:
\begin{eqnarray}
&&y_{ij}^k \thicksim \mathcal{N} (c_k d_{ij}^k+h_k, \sigma_k^2), \label{eq:target}
\end{eqnarray}
where $c_k, h_k$, and $\sigma_k \in \Re$ are universal (i.e., not subject-specific)  parameters associated with the $k$'th target variable.
We refer to Eq.~(\ref{eq:target}) as an observation model.
Fig.~\ref{Fig:dependency} depicts the dependency relationship between all variables.
\subsection{Inference}
In this section, we discuss how to train the proposed model and apply it during test time.

\subsubsection{Training}

Let us use $\mathbf{\Theta}$ to denote the parameter set of our model:
 \begin{eqnarray}
\nonumber &&\mathbf{\Theta} = \{ \w, b, \sigma_p , \sigma_s, \v, \{a_k, c_k, h_k, \sigma_k \}_{k=1,\cdots,m}\}.
\end{eqnarray}

The goal of training is to estimate the model parameters $\mathbf{\Theta}$ given data from $n$ subjects: $\{ \y_i, \x_i, \t_i \}_{i=1, \ldots, n}$.
Here, $\y_{i} = [\y_i^1 \ldots \y_i^m] \in \Re^{v_i \times m}$ denotes $m$ target values of the $i$th subject for $v_i$ visits.
We estimate $\mathbf{\Theta}$ via maximizing the likelihood function :
\begin{eqnarray}
\nonumber&& \prod_{i=1}^n P(\y_{i}| \x_i, \t_i; \mathbf{\Theta}).
\end{eqnarray}
Note that we use the standard notation of $p(y|x)$ to indicate the probability density function of the random variable $Y$ (evaluated at $y$) conditioned on the random variable $X$ taking on the value $x$.
Also, parameters not treated as random variables are collected on the right hand side of ``$;$''.

Now, let us focus on the likelihood of each subject:
\begin{eqnarray}
\label{Eq:PXY2}
&&P(\y_{i}| \x_i, \t_i; \mathbf{\Theta}) \label{eq:likelihood} \\
&&= \int \int \left[ \prod_{j=1}^{v_i} p(\y_{ij} | s_i, \p_i, \t_{ij}) \right] p(s_i, \p_i | \x_i; \mathbf{\Theta}) ds_i d\p_i, \nonumber
\end{eqnarray}
with $p(s_i, \p_i | \x_i; \mathbf{\Theta}) = p(s_i | \x_i; \mathbf{\Theta}) p(\p_i | \x_i; \mathbf{\Theta})$ due to Eq~(\ref{eq:p},\ref{eq:s}).

Instead of the computationally challenging Eq~(\ref{eq:likelihood}), we use variational approximation~\cite{ranganath2014black} and maximize the expected lower bound objective (ELBO):
\begin{eqnarray}
F( \mathbf{\Theta}, \{\mathbf{\gamma}_i\}) &=&\sum_{i=1}^{n} \mathbb{E}_q (\sum_{j=1}^{v_i} \sum_{k=1}^m \log p(y_{ij}^k | s_i, p_i^k, \t_{ij}; \mathbf{\Theta})) \nonumber \\
&-& \mathbb{E}_q (\log q(s_i; \mathbf{\gamma}_i) )-  \mathbb{E}_q (\log q(\p_i; \mathbf{\gamma}_i)), \label{eq:ELBO}
\end{eqnarray}
where $q(s_i; \mathbf{\gamma}_i) = N(\mu_{si}, \sigma_{si}^2)$ and $q(\p_i; \mathbf{\gamma}_i)) = N(\mathbf{\mu}_{pi}, \mathbf{\Sigma}_{pi} = \Gamma_{pi}^T \Gamma_{pi})$ are proxy distributions that approximate the true posterior distributions $p(s_i |\y_i,  \x_i; \mathbf{\Theta})$ and  $p(\p_i |\y_i,  \x_i; \mathbf{\Theta})$, respectively.
Recall that $\p_i$ is $m$-dimensional since each target variable is associated with a different inflection point, yet the slope parameter $s_i$ is shared across targets and thus a scalar.
We have used $\mathbf{\gamma}_i =  \{\mu_{si}, \sigma_{si}, \mathbf{\mu}_{pi}, \Gamma_{pi} \}$ to collectively denote the proxy parameters.
The expectation in the first term is with respect to the proxy distributions and can be approximated via Monte Carlo sampling. Thus:
\begin{eqnarray}
&& \mathbb{E}_q (\sum_{k} \log p(y_{ij}^k | s_i, p_i^k, \t_{ij}; \mathbf{\Theta})) \nonumber \\
&& \approx \frac{1}{S} \sum_{s = 1}^S \log p(\y_{ij} | s_i^{(s)}, \p_i^{(s)}, \t_{ij}; \mathbf{\Theta}),\label{eq:MC}
\end{eqnarray}
where $s_i^{(s)}$ and  $\p_i^{(s)}$ are Monte Carlo samples drawn using the ``re-parameterization trick.''
I.e., $s_i^{(s)} = \eta^{(s)} \sigma_{si} +  \mu_{si}$ and $\p_i^{(s)} =  \Gamma_{pi}^T \mathbf{\epsilon}^{(s)} + \mathbf{\mu}_{pi}$,
where $\eta^{(s)} \in \R$ and $\mathbf{\epsilon}^{(s)} \in \R^{m \times 1}$ are realizations of the auxiliary random variables, independently drawn from zero-mean standard Gaussians, $N(0, 1)$ and $N(\mathbf{0}, \mathbf{I})$, respectively.
The ``re-parameterization trick'' allows us to differentiate the ELBO (or more accurately, its approximation that uses Eq.~\ref{eq:MC}) with respect to $\mathbf{\gamma}_i$.
%
E.g.:
\begin{eqnarray}
\frac{\partial s_i^{(s)}}{\partial \sigma_{si}} = \eta^{(s)}, \text{ and } \frac{\partial s_i^{(s)}}{\partial \mu_{si}} = 1. \nonumber
\end{eqnarray}

During training, we use gradient-ascent to iteratively optimize Eq.~\ref{eq:ELBO} with respect to $\mathbf{\Theta}$ and the parameters of the proxy distributions: $\{\mathbf{\gamma}_i \}$.
Note the structure of ELBO is flexible: missing target variables are treated by ignoring the corresponding term and the sum is over visits, which can handle irregular timings.
Training yields optimal parameters: $\mathbf{\Theta^*}$, and $\{ \gamma_{i}^* \}$.

\subsubsection{Testing}

During test time, we are interested in computing the posterior distribution of $\y_{n+1}$ for a new subject with $\x_{n+1}$ at an arbitrary time-point (age) $t$.
Note that we drop the second sub-script, i.e., $j$ index, of $\y_{n+1}$ to emphasize that we will be computing these posterior probabilities at many different (often future) time-points.
There are two types of test subjects: those with no history of visits (scenario 1), and those with at least one prior clinical visit (scenario 2).
For scenario 2, we will use $\{ \y_{(n+1)j}, t_{(n+1)j} \}_{j=1, \ldots, v_{n+1}}$ to collectively denote the $v_{n+1}$ historical observations and their corresponding visit times.
Note that we fix $\mathbf{\Theta^*}$ to the values obtained from training.

In scenario 1, we use Eq.~(\ref{eq:likelihood}) to compute the posterior:
\begin{eqnarray}
&&P(y_{n+1}^k | \x_{n+1}, t; \mathbf{\Theta}^*) \label{eq:likelihood2} \\
&&= \int \int p(y_{n+1}^k | s, p^k, t; \mathbf{\Theta}^*)  p(s | \x_{n+1}; \mathbf{\Theta}^*) p(p^k| x_{n+1}; \mathbf{\Theta}^*) ds d p^k, \nonumber
\end{eqnarray}
where $p(s | \x_i; \mathbf{\Theta}^*)$ and $p(p^k| x_i; \mathbf{\Theta}^*)$ are defined in Eq~(\ref{eq:p},\ref{eq:s}).

In the second scenario, we will first maximize the ELBO of Equation~(\ref{eq:ELBO}) evaluated for the observations on the new subject $\{ \y_{(n+1)j}, t_{(n+1)j}\}$ and attribute vector: $\x_{n+1}$:
\begin{eqnarray}
\mathbb{E}_q (\sum_{j=1}^{v_{n+1}} \sum_{k=1}^m \log p(y_{(n+1)j}^k | s_{n+1}, p_{n+1}^k, \t_{(n+1)j}; \mathbf{\Theta}^*) \nonumber \\
- \mathbb{E}_q (\log q(s_{n+1}, \mathbf{\gamma}_{n+1})) -  \mathbb{E}_q (\log q(\p_{n+1}, \mathbf{\gamma}_{n+1})). \label{eq:ELBO2}
\end{eqnarray}
Eq~(\ref{eq:ELBO2}) is optimized with respect to $\mathbf{\gamma_{n+1}}$, which yields proxy distributions that can be viewed as approximations:
\begin{eqnarray}
q(s_{n+1}, \mathbf{\gamma}_{n+1}) &\approx& p(s | \{\y_{(n+1)j}, t_{(n+1)j} \}, \x_{n+1}; \mathbf{\Theta}^*) \label{eq:p2} \\
q(\p_{n+1}, \mathbf{\gamma}_{n+1}) &\approx& p(p^k| \{\y_{(n+1)j}, t_{(n+1)j} \}; \mathbf{\Theta}^*) \label{eq:s2}
\end{eqnarray}
Note that these distributions can be regarded as a customization of the priors on $s$ and $\p$ given the observed data.
We then proceed to use these approximate $q$ distributions in Equation~(\ref{eq:likelihood2}), replacing $p(s | \x_i; \mathbf{\Theta}^*)$ and $p(p^k| x_i; \mathbf{\Theta}^*)$, to evaluate the posterior distribution for an arbitrary time-point $t$ conditioned on past observations.

During test-time, we often have two distinct objectives: maximizing the posterior distribution or drawing samples from it to estimate the posterior mean and standard deviation.
For the maximization problem, we can approximate the integral of Eq~(\ref{eq:likelihood2}) via a Monte Carlo strategy by drawing samples from $p(s | \x_i; \mathbf{\Theta}^*)$ and $p(p^k| x_i; \mathbf{\Theta}^*)$ (Scenario 1) or the approximate distributions $q(s_{n+1}, \mathbf{\gamma}_{n+1})$ and $q(\p_{n+1}, \mathbf{\gamma}_{n+1})$ (Scenario 2).
Finally, one can use an ancestral sampling strategy to generate samples from the posterior distribution.
Here, we first sample $s_{n+1}$ and $\p_{n+1}$, either from the priors of Eq~(\ref{eq:p}) and (\ref{eq:s}) (Scenario 1) or customized priors of Eq~(\ref{eq:p2}) and (\ref{eq:s2}) (Scenario 2).

\section{Experiments}
\subsection{Dataset}

We use a dataset of 3,057 subjects (baseline age $73.3 \pm 17.2$ years) collected by ADNI~\cite{adni} to empirically validate and demonstrate the proposed model.
This dataset contained multiple clinical visits per subject, during which thorough cognitive and symptomatic assessments were conducted.
In our experiments, we used MMSE, ADAS-COG and CDR-SB as three target variables.
MMSE has a range between 0 (impaired) and 30 (healthy), whereas ADAS-COG takes on values between  0 (healthy) to 70 (severe), and CDR-SB varies from 0 (healthy) to 18 (severe).
The first two (MMSE and ADAS-COG) are general cognitive assessments that track and predict dementia, while CDR-SB is a clinical score that measures the severity of dementia-associated symptoms.

In ADNI, clinical assessment were done every 6-12 months.
The timing of these visits varied and certain subjects missed visits.
Furthermore, most subjects dropped out of the study by their 4th planned visit.
Hence while the clinical follow-up period spanned over 5 years, each subject had an average of $3.2$ visits.
In total, there were $9716$ time-points in our dataset.
The subjects ranged from $55$ to $95$ years of age and were grouped into three clinical categories: health control (HC), mild cognitive impairment (MCI), and AD patients.
These clinical categorizations in part relied on the target variables of interest.
Table~\ref{Tb:demo} provides a summary of the different visits and how they breakdown across clinical groups.
During the follow-up period, some subjects transitioned between categories, resulting in six types of subjects: stable HC, stable MCI,  stable AD, and MCI-to-AD, HC-to-AD, and HC-to-MCI converters. There were also a very small number of subjects who improved in clinical categories (e.g. AD-to-MCI).

In addition to the target variables, we utilized individual-level traits associated with AD: age, APOE genotype (number of E4 alleles)~\cite{corder1993gene}, sex, and education (in years)~\cite{fratiglioni1991prevalence}.
We also used baseline brain MRI scans to derive two anatomical biomarkers of AD: total hippocampal and ventricle volume normalized by brain size.
These imaging biomarkers were automatically computed with FreeSurfer~\cite{fischl2012freesurfer} and quality controlled as previously described~\cite{mormino2016polygenic}.

Finally, as we describe below, we considered utilizing the longitudinal imaging biomarkers (hippocampal and ventricle volume) as target variables that were available during training.
This is because ADNI is a unique dataset and attempts to acquire brain MRI scans on each subject every 6-12 months.
As a result, we have access to these invaluable data.
Yet, we emphasize that during test time, we only considered the availability of MMSE, ADAS-COG and CDR-SB on historical visits and did not assume longitudinal neuroimaging data on test subjects.
\begin{table}[]
\centering
\caption{Number of subjects per baseline clinical group. The time interval (mean $\pm$ std) between subsequent visits.\label{Tb:demo}}

\begin{tabular}{|c|*4c|}
\hline
Visit &  \multicolumn{3}{c}{Number of Subjects} & Time From \\
Number &  HC & MCI & AD &  Prior Visit (mo) \\
\hline
baseline & 797& 1709 & 551 & N/A \\
1&  611 & 1070&312 & $7.4\pm 21.4$ \\
2 &  450 &758 &194 & $8.0\pm 19.8$ \\
3& 338 & 486 &92 &$10.1\pm 19.2$ \\
4& 241 &331 & 11&$10.5\pm 18.6$ \\
5&  155& 189&0 &$11.2\pm 12.7$ \\
6 & 92 &120 &0 &$10.3\pm 13.2$ \\
7 &45 & 59& 0&$8.9\pm 7.8$\\
\hline
\end{tabular}
\end{table}

\subsection{Experimental Setup}
\subsubsection{Benchmark Methods}

In our experiments, we compare the proposed method to the following benchmarks:
\begin{enumerate}
\item Global: A 4-parameter (scale, bias, inflection, and slope) sigmoidal model that was fit on all training data (least-squares)
\item Sex-specific: Same as above but separate for males and females
\item APOE-specific: Same as above, but separate for three groups defined by APOE-E4 allele count $\{0,1,2\}$.
\item Sex- and APOE-specific: Same as above, but separate for each sex and APOE group.
\item Linear mixed effects (LME) model: A linear regression model with subject-specific attributes ($\x_i$) as fixed effects, and time and bias term as a random effects.
This LME model, commonly used to capture longitudinal dynamics~\cite{linearrandom,bernal2013}, allows each subject to deviate from the average trajectory determined by its attributes by shifts in slope and offset.
\item Subject-specific linear model: Least-squares fit of a linear model on each subject's historical data.
When there is only one past visit, we adopt a carry-forward extrapolation.
\end{enumerate}

Benchmarks 1-5 make use of the training data, whereas benchmark 6 ignores the training data and merely relies on each test subject's own data.
While benchmark 1-4 use models that are fixed after training, benchmarks 5 and 6 make adaptations to the model given observations on the test subject.
Below, we refer to benchmarks 1-4 as training-fixed benchmarks.
The LME model (benchmark 5) uses test observations to estimate subject-specific deviations from a global linear model.
Benchmark 6 fits a line to historical data of the test subject, and is a widely used technique in clinical practice. We did not fit a sigmoid to the subject-specific data as one would need more than 4 historical time-points to obtain reliable estimates.
For benchmarks 1-4, we also implemented linear versions (i.e. least squares fit of a line), yet the prediction performance was no better than the sigmoidal modes.
Therefore, we omitted those results due to space constraints.

\subsubsection{Evaluation}
For each target variable, we use the mean and standard deviation of the absolute error across test subjects to evaluate the different models.
In order to examine the statistical significance of the difference between the proposed method and benchmarks, we used following ``paired permutation'' strategy.
For each test subject, we randomly permuted the labels of the prediction models.
Note that  the random permutation was done at the subject-level, in order to respect the temporal dependency structure in longitudinal assessments.
Thus, each randomly shuffled model was used to compute predictions for all time-points of a given subject.
For each permutation, we computed the mean absolute error (MAE, average across target variables) of the predictions of each (randomly shuffled) model.
Next, we computed and saved the difference between each model's MAE and the proposed model's MAE.
After sorting these differences (in descending order) over all permutations, the permutation p-value was computed as the rank of the true difference (i.e. the difference between the MAE of the benchmark model and proposed model without permutation) divided by the number of permutations (10,000) plus 1.

\subsubsection{Implementation Details}
We implemented the proposed model and inference algorithm in Python~\footnote{The code of this work is available at https://github.com/zyy123jy/kdd}, using the Edward library~\cite{tran2016edward}, which is in turn built on TensorFlow~\cite{abadi2016tensorflow}. We used a 20-fold cross-validation strategy in all our experiments.
We first partitioned the data into 20 non-overlapping, roughly equally-sized sets of subjects.
In each of the 20 folds, we reserved one of the partitions as the independent test set.
Out of the remaining 19 partitions, one was set aside as a validation set, while the rest were combined into a training set.
The training set was used to estimate the model parameters, i.e., $\mathbf{\Theta^*}$, while performance on the validation set was used to select hyper-parameters, such as step size in the optimization and evaluate random initializations.
Finally, test performance was computed on the test set.
We report results averaged across 20 folds.


\subsection{Prediction Results}

We first show the quantitative prediction results for all the methods and three target variables (MMSE, ADAS-COG, and CDRSB).
In the following, we consider several prediction scenarios.

In the first prediction scenario, we vary the number of past visits available on the test subjects (i.e., $v_{n+1}$).
In general, we expect this variation to influence the LME and subject-specific linear model benchmarks (5 and 6), in addition to the proposed model.
These methods fine-tune their predictions based on historical observations available on test data.
With more test observations, we expect them to achieve better accuracy.
All other benchmarks are fixed after training and thus their performance should not improve with increasing number of past observations.

In the second scenario, we fix the number of past observations on test subjects and vary the prediction horizon.
In general, all models' predictions should be less accurate for more distant future time-points.

Finally, we focus on the proposed model and consider training it on longitudinal imaging data available in training.
Brain MRI scans are expensive and hard to obtain, so throughout this paper we assumed that each test subject has only a single baseline MRI scan.
Yet, the ADNI dataset contains longitudinal imaging data and we were interested in quantifying the effect of using these during training on test-time performance.


\subsubsection{Varying the Number of Past Visits}
\begin{figure*}[t]
\centering
\includegraphics[width=15.5cm]{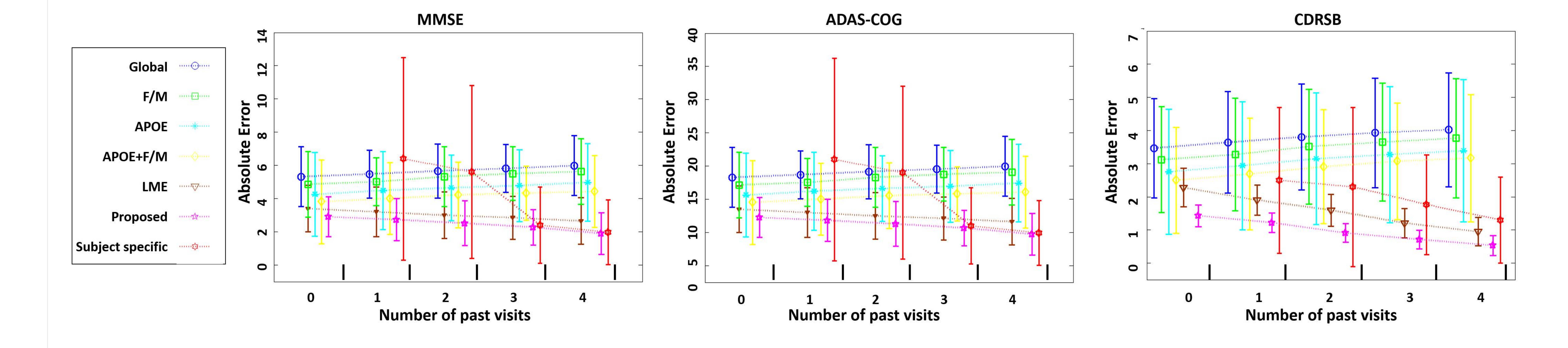}
\caption{Absolute error (mean and standard derivation) of all methods for predicting MMSE, ADAS-COG and CDRSB, as a function of number of past visits available on test subjects.
\label{Fig:mmsr-abs}}
\end{figure*}

Figure~\ref{Fig:mmsr-abs} shows the MMSE, ADAS-COG and CDRSB prediction accuracies (mean and standard deviation of absolute error).
We observe that the performance of the training-fixed benchmarks (1-4) worsen slightly as the number of past visits increases.
This is likely because the training data contains more samples at early times (i.e., relatively younger ages), partially because most subjects drop out by their 4th visit.
Therefore, a model trained on these data is expected to be less accurate for older ages.

The adaptive benchmarks (5-6) and the proposed model, on the other hand, overcome this handicap to achieve better accuracy with more past visits.
As we discussed above, this is largely because these techniques exploit test observations to fine-tune their models.
The subject-level linear model (benchmark 6), in fact, is an extreme example, where the predictions are computed merely by extrapolating from historical observations without relying on training data.

Finally, we note that the proposed model achieves a significantly and substantially better accuracy than all benchmarks (all paired permutation p-values $< p_{max}=0.04$).
The subject-specific benchmark (6) exhibits the largest variance implying the quality of performance varies wildly across subjects.
Overall, the training-fixed benchmarks perform the worst.
In general the proposed model's variance is among the smallest, indicating consistency in prediction accuracy.
%

\subsubsection{Varying the Time Horizon}

In order to evaluate how prediction performance changes as a function of the time horizon, we evaluated the methods for different future time-points.
In this empirical scenario, we assume that each test subject has 2 past clinical assessments (obtained at baseline and month 6).
Our goal is to predict  MMSE, ADAS-COG and CDRSB scores at later time-points (starting at 12 months after baseline, up to 36 months).
Based on the longitudinal study protocol, we considered 6 month intervals and assigned the actual visits to the closest 6-month bucket.

\begin{figure*}[t]
\centering
\includegraphics[width=15.5cm]{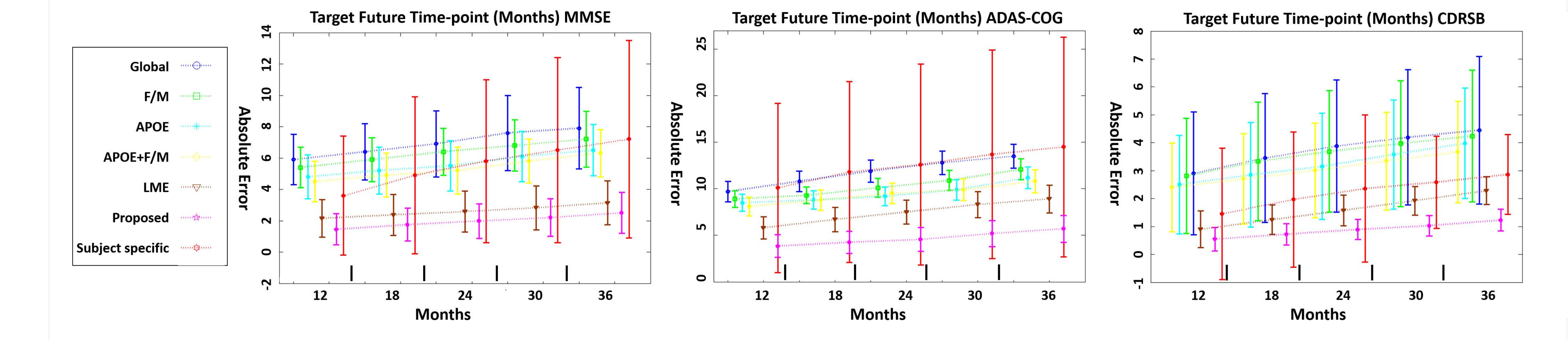}
\caption{Absolute error (mean and standard derivation) of all methods for predicting MMSE, ADAS-COG and CDRSB.
We used two points from each test subject as past observations and varied the time horizon for prediction. \label{Fig:mmsr-abs3}
}
\end{figure*}

Fig~\ref{Fig:mmsr-abs3} shows prediction accuracies of all considered methods.
The proposed method performs significantly (all paired permutation p-values $ < p_{max}=0.03$) and substantially better than all other methods, with the difference increasing from the short term (12 months) to long term (36 months).
For the benchmark models, prediction accuracy tends to drop more dramatically for longer time horizons.
As above, training-fixed benchmarks perform the worst.


\subsubsection{Training  with or without longitudinal MRI scans}
 Since longitudinal MRI scans are available for most subjects in our dataset, we considered the use of serial imaging biomarkers in training.
 Yet, as before, we assume that only the baseline MRI scan is available for testing subjects.

 Adding a time-varying biomarker is relatively easy in our model, as it involves adding another target variable type.
 This target variable will have its own inflection point and observation model (Equation~\ref{eq:target}).
 Note that in our framework, due to the conditional independence assumptions, inference is robust to the timing and availability of the target variables.
 During training (i.e., when optimizing the ELBO function), the algorithm simply sums over all available visits and evaluates the function only for observed targets.

\begin{figure*}[t]
\centering
\includegraphics[width=14cm]{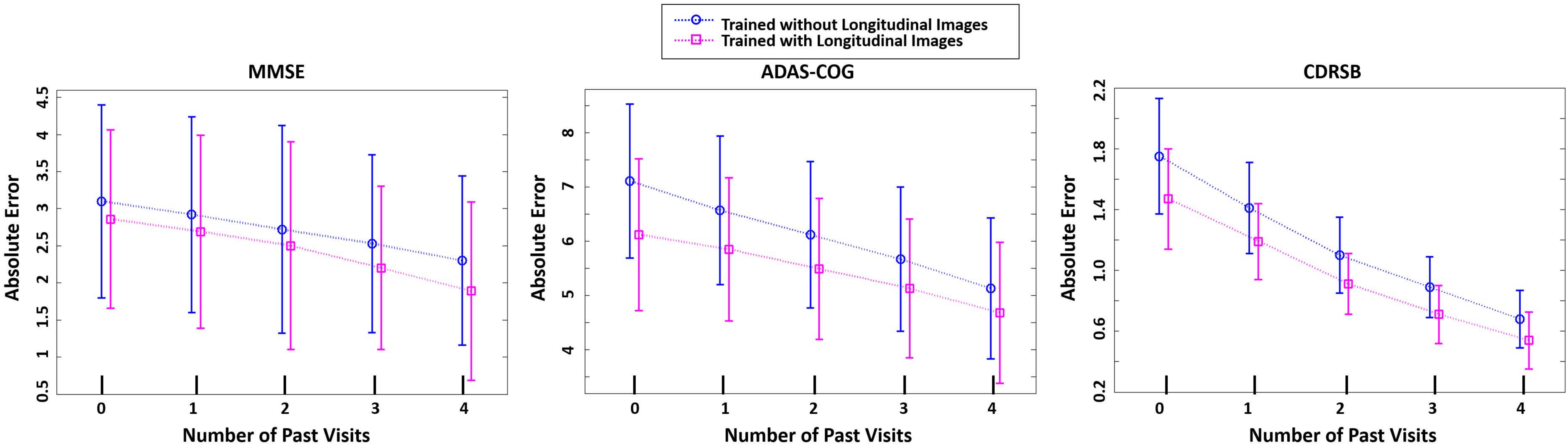}
\caption{The mean absolute errors with standard derivation of our method on MMSE prediction computed from 20-fold cross-validation with longitudinal MR images in training process and without. We varies the number of past visits from 0 to 4 to evaluate how our
 methods adapt to use different priors from past visits to improve the performance.  \label{Fig:mmsr-abs2}}
\end{figure*}

Fig. \ref{Fig:mmsr-abs2} shows the performance of our model trained with and without longitudinal MRI scans.
We observe that prediction accuracy is consistently better when trained with longitudinal MRI scans, which suggests that we can improve the quality of the model's predictions by incorporating additional time-varying biomarkers.

\subsection{Further Analysis of the Proposed Model}
The proposed model can be viewed as a generative probabilistic model that gives rise to clinical assessments and possibly other time-varying biomarkers.
Hence, in addition to computing individual-level predictions, we can probe the trained model to gain further insights about the underlying dynamics of the pathology.
In this section, we provide examples of such analyses.

\subsubsection{Inflection Points}
In our model, each target variable is associated with a latent sigmoid curve that captures temporal dynamics.
While the slope of these sigmoids are coupled across target variables, their inflection points $p^k$ are different.
Yet, we emphasize that the inflection points are not assumed to be independent across targets, as they are drawn from a prior distribution shaped by the subject's attributes $\x_i$ (see Eq.~\ref{eq:p}).
For a given subject, the difference between the inflection points of a pair of target variables reflects the time lag between the corresponding progression curves.

\begin{figure}[h]
\centering
\includegraphics[width=8cm]{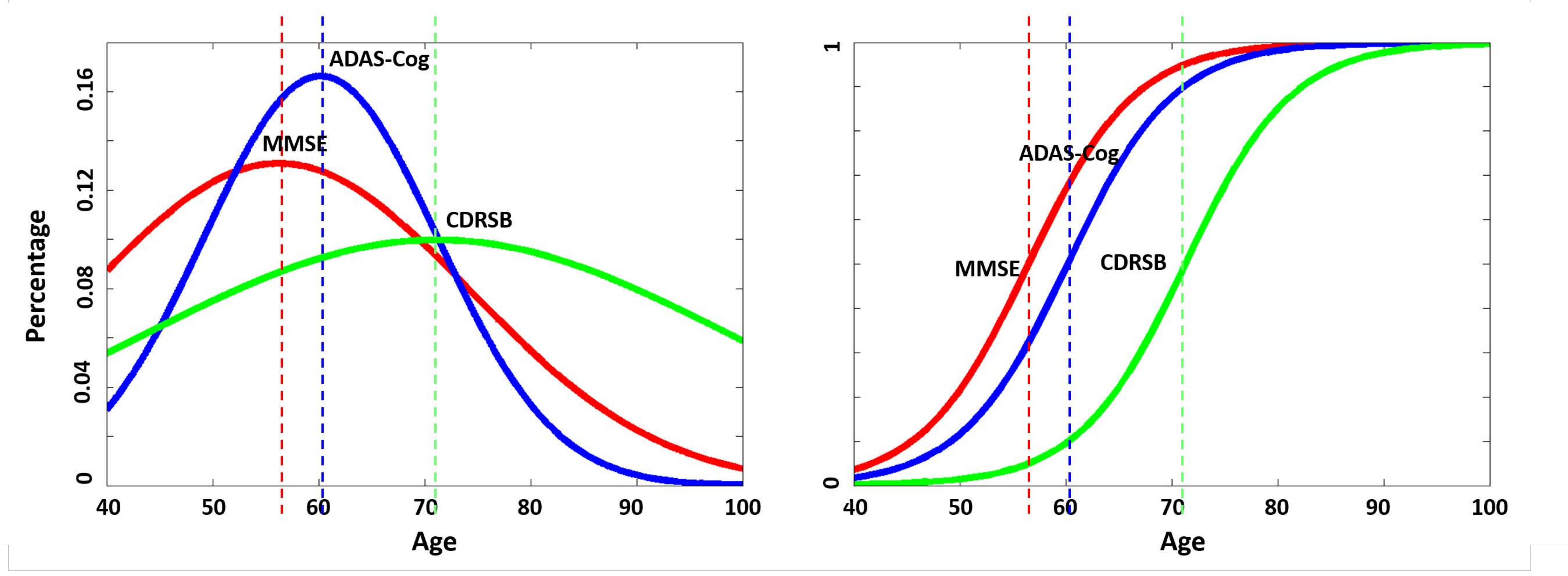}
\caption{Left: empirically estimated prior distributions of the inflection points for the three clinical target variables. Right: Mean latent progression curves for the different targets.\label{Fig:inf}}
\end{figure}

After we fit the proposed model on the training data, we draw samples for the unobserved inflection points on test data.
We then empirically estimated the prior distribution of $p^k$ averaged across subjects via a kernel density estimator (Gaussian kernel with variance = 2.5).
Fig.~\ref{Fig:inf} plots the empirical prior distributions for the different target variables and their latent progression curves corresponding to the mean parameter values (for the slope and inflection points) derived from the empirical priors.

We notice that the MMSE progression, on average, is earlier than ADAS-COG, with a mean difference of around 3.5 years.
ADAS-COG is, in turn, on average, about 11 years earlier than CDR-SB.
These results are consistent with the hypothesized trajectories of AD biomarkers~\cite{Jack2010}, where memory and cognitive scores such as MMSE and ADAS-COG start declining sooner than clinical symptoms.

%

\subsubsection{The Impact of APOE, Sex, and Education}

Next, we were interested in inspecting how APOE, sex, and education impact the trajectories MMSE, ADAS-COG and CDRSB.
In our model, the parameter vector $\w$ determines the mean slope and $\v$ affects the mean inflection point of the latent progression curves. 
Table~\ref{Tb:weights} lists the estimated values in $\w$ and $\v$ (averaged over the 20 folds) for APOE, sex, and education.
We observe that each extra APOE E4 allele copy increases the maximum rate of progression (which is equal to $s_i/4$) by an additional $24.5 \%$, yet shifts the inflection forward by $0.11$ years.
Males, on average, have a maximum rate of progression that is $9.25 \%$ more than females, and their inflection points are $0.31$ years later.
Each additional decade of education, on the other hand, delays the progression curves by $0.44$ years, yet the maximum rate of progression increases by an additional $8.1\%$.

\begin{table}[h]
\centering
\caption{Average estimates for $\w$ and $\v$ corresponding to APOE, sex, and education}
\label{Tb:weights}
\begin{tabular}{|l||l|l|l|}
\hline
 & APOE & Sex  & Edu \\
 & (E4 Count) & (0: female, 1: male) & (decades) \\
\hline
 $\w$& $0.98 $ &    $0.38 $ &   $0.33$ \\
 $\v$ &  $0.11$ &   $0.31 $&   $ 0.44$ \\
\hline
\end{tabular}
\end{table}
%

%
%
\subsubsection{Visualization of Personalized Models}
\begin{figure*}[h]
\centering
\includegraphics[width=16cm]{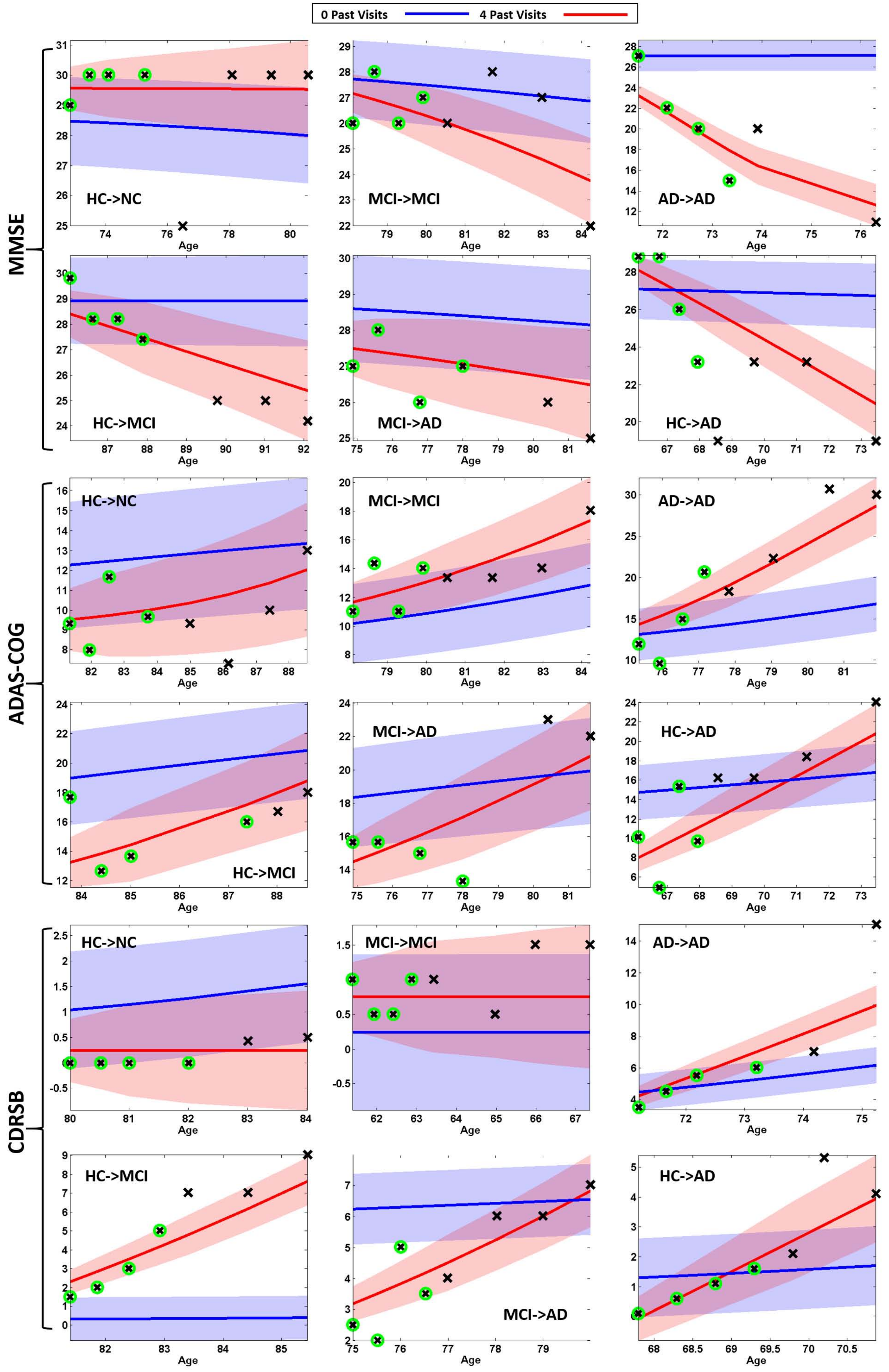}
\caption{Visualization of predicted (solid line, posterior mean) MMSE, ADAS-COG and CDRSB curves for various representative subjects. One standard deviation of predictions are illustrated with shaded area ($68.2 \%$ confidence interval). Clinical assessments are indictated with \textbf{x}. The proposed model was used under two conditions: with no prior visits (blue, baseline) and with the first four visits (green circles) treated as test observations (red, personalized). The ground truth future visits are denoted by orange rectangles. See text for further details. }
\label{Fig:cdrsb-exam}
\end{figure*}
As we discussed above, the proposed model can adjust its predictions based on available observations on the test data.
In this section, we were interested in revealing this personalization effect on subject-specific trajectories.
In Figure~\ref{Fig:cdrsb-exam}, we visualized the predicted trajectories for MMSE, ADAS-COG and CDRSB on several representative subjects, under two conditions: no past visits and four past visits.
Note that we chose the subjects from 6 sub-groups: stable HC, stable MCI, stable AD, HC-to-MCI converter, MCI-to-AD converter, and HC-to-AD converter.
In each sub-group, we selected representative subjects with most time-points.
Figure~\ref{Fig:cdrsb-exam} visualizes the ground truth clinical scores, which are marked with $x$'s.
Those points that are circled with green are considered observed past points in the second condition to compute the adjusted projections (personalized model).
We observe that the baseline model predictions (corresponding to no past visit and illustrated in blue) are often much less accurate than the personalized model (red).
The baseline model, in general, predicts little change over time, whereas the personalized model offers better projections that can capture significant change.

\section{Discussion}

We presented a probabilistic, latent disease progression model for capturing the dynamics of the underlying pathology that is often shaped by risk factors such as genotype.
Our work is motived by real-world clinical applications, where irregular visiting patterns, missing variables, and inconsistent multi-modal assessments are ubiquitous.
In the proposed framework, we make a distinction between subject-level attributes, which we assume are fixed, and time-varying clinical observations, which can include imaging and cognitive tests, collected over multiple visits.
These time-varying variables are modeled to be noisy observations of an idealized latent, sigmoid progression curve that has two parameters: its inflection point and slope.
These parameters are in turn, assumed to be functions of the fixed subject attributes.
We take a Bayesian approach to fit this model on a training dataset with clinical observations, where the parameters of the latent progression curve is integrated out.

In this work, we had two distinct, yet simultaneous goals.
Our first goal was practical: to forecast the clinical future of a test subject based on a population model that is customized if we have historical observations from the test subject.
Our second goal was ``knowledge discovery:'' we were interested in gaining insights about the underlying dynamics of various clinical assessments and further identifying the impact of risk factors such as genotype.

We applied the proposed method on a large dataset of Alzheimer's disease with promising results.
In our experiments, we analyzed prediction accuracy for the proposed model and several benchmark methods under different scenarios that included varying the known patient history and prediction horizon.
In all our comparisons, the proposed model achieved the best performance.
However, we caution the author to treat these results as preliminary.
In future work, we are interested in exploring methods that might yield better prediction accuracy.
One approach would be to exploit more input variables, such as genome-wide markers and whole-brain images.
Such high-dimensional models will likely require additional prior constraints, such as sparsity, to avoid overfitting, as in~\cite{ADkdd11,Xiang2013}.
Another, closely related framework is multi-task learning, which is suitable for our problem of predicting multiple clinical target variables.
Multi-task learning has been applied to predict multiple correlated diseases~\cite{Nori2015}, integrate data from several sources~\cite{dg2015} or modalities~\cite{daoqiangzhang2011}.
However, in prior works, the target variables are often considered fixed and/or longitudinal observations are ignored.

Another direction that will likely yield improved accuracy is to extend our model beyond linearity assumptions.
Deep learning techniques, such as reccurent neural networks (RNNs), offer a natural framework for this direction.
RNNs have been applied to related problems, such as clinical diagnosis from time-series data~\cite{Lipton2015LearningTD, RNN2016}, and imputing missing longitudinal variables~\cite{RNNmissingdata}.
The main challenge in these approaches is that they are often data hungry (requiring lots of annotated data to train on) and hard to interpret, which conflicts with our objective to obtain interpretable insights about underlying disease dynamics.
Furthermore, RNN models usually assume fixed intervals between time-points - an assumption widely violated in real-life scenarios.
In contrast, our Bayesian approach is flexible, assumes no regularity in visit times, does not need a huge amount of data to fit, can seamlessly handle multiple modalities, and, importantly, yields straightforward interpretations.

\section{Acknowledgements}
This work was supported by National Institutes of Health (NIH) grants  R01LM012719, R01AG053949, and 1R21AG050122, and the National Science Foundation NeuroNex Neurotechnology Hub grant 1707312.

In our experiments, we employed data downloaded from the longitudinal ADNI study (phases 1, GO, and 2, adni.loni.usc.edu)~\cite{adni}, based on the data derived for the Tadpole 2017 Challenge (https://tadpole.grand-challenge.org/home/).
Scripts to generate these data from ADNI can found here: \\
https://github.com/noxtoby/TADPOLE.

The ADNI was launched in
2003 as a public-private partnership, led by Principal Investigator Michael W. Weiner,
MD. The primary goal of ADNI has been to test whether serial magnetic resonance imaging
(MRI), positron emission tomography (PET), other biological markers, and clinical and
neuropsychological assessment can be combined to measure the progression of mild
cognitive impairment (MCI) and early AlzheimerÕs disease (AD). For up-to-date information,
see www.adni-info.org.

\bibliographystyle{abbrv}
\bibliography{sigproc1}

\begin{thebibliography}{10}

\bibitem{ADearly}
The need for early detection and treatment in alzheimer's disease.
\newblock {\em EBioMedicine}, pages 1--2, 2018.

\bibitem{abadi2016tensorflow}
M.~Abadi et~al.
\newblock {TensorFlow:} a system for large-scale machine learning.
\newblock In {\em OSDI}, 2016.

\bibitem{AD2010}
A.~Association.
\newblock Alzheimer’s disease facts and figures.
\newblock {\em Alzheimer’s and Dementia}, pages 158--194, 2010.

\bibitem{bernal2013}
J.~Bernal-Rusiel et~al.
\newblock Statistical analysis of longitudinal neuroimage data with linear
  mixed effects models.
\newblock {\em NeuroImage}, 2013.

\bibitem{cano2010adas}
S.~J. Cano et~al.
\newblock The {ADAS-COG} in {Alzheimer`s} disease clinical trials: psychometric
  evaluation of the sum and its parts.
\newblock {\em Journal of Neurology, Neurosurgery \& Psychiatry}, 2010.

\bibitem{RNN2016}
E.~Choi et~al.
\newblock Doctor {AI}: Predicting clinical events via recurrent neural
  networks, 2016.

\bibitem{corder1993gene}
E.~H. Corder et~al.
\newblock Gene dose of {Apolipoprotein E} type 4 allele and the risk of
  alzheimer's disease in late onset families.
\newblock {\em Science}, 261(5123):921--923, 1993.

\bibitem{fischl2012freesurfer}
B.~Fischl.
\newblock Freesurfer.
\newblock {\em Neuroimage}, 2012.

\bibitem{fratiglioni1991prevalence}
L.~Fratiglioni et~al.
\newblock Prevalence of {Alzheimer's disease} and other dementias in an elderly
  urban population relationship with age, sex, and education.
\newblock {\em Neurology}, 1991.

\bibitem{jack1999prediction}
C.~Jack et~al.
\newblock Prediction of {AD with MRI-based} hippocampal volume in mild
  cognitive impairment.
\newblock {\em Neurology}, 1999.

\bibitem{Jack2010}
C.~Jack et~al.
\newblock Hypothetical model of dynamic biomarkers of the {Alzheimer's}
  pathological cascade.
\newblock {\em The Lancet Neurology}, 2010.

\bibitem{katzman1993education}
R.~Katzman.
\newblock Education and the prevalence of dementia and alzheimer's disease.
\newblock {\em Neurology}, 1993.

\bibitem{RNNmissingdata}
H.~G. Kim et~al.
\newblock Recurrent neural networks with missing information imputation for
  medical examination data prediction.
\newblock {\em 2017 IEEE International Conference on Big Data and Smart
  Computing (BigComp)}, pages 317--323, 2017.

\bibitem{Lipton2015LearningTD}
Z.~Lipton et~al.
\newblock Learning to diagnose with lstm recurrent neural networks.
\newblock {\em CoRR}, abs/1511.03677, 2015.

\bibitem{cancer2013}
E.~G. Luebeck et~al.
\newblock Impact of tumor progression on cancer incidence curves.
\newblock pages 1086--1096, 2012.

\bibitem{mmse}
M.~Marta.
\newblock Modelling mini mental state examination changes in alzheimer's
  disease.
\newblock 2000.

\bibitem{linearrandom}
C.~McCulloch et~al.
\newblock Prediction of random effects in linear and generalized linear models
  under model misspecification.
\newblock {\em Biometrics}, pages 270--279, 2010.

\bibitem{moradi2015machine}
E.~Moradi et~al.
\newblock Machine learning framework for early {MRI-based Alzheimer's
  conversion prediction in MCI subjects}.
\newblock {\em Neuroimage}, 2015.

\bibitem{mormino2016polygenic}
E.~Mormino et~al.
\newblock Polygenic risk of alzheimer disease is associated with early-and
  late-life processes.
\newblock {\em Neurology}, 2016.

\bibitem{nestor2008ventricular}
S.~Nestor et~al.
\newblock Ventricular enlargement as a possible measure of alzheimer's disease
  progression validated using the alzheimer's disease neuroimaging initiative
  database.
\newblock {\em Brain}, 2008.

\bibitem{Nori2015}
N.~Nori et~al.
\newblock Simultaneous modeling of multiple diseases for mortality prediction
  in acute hospital care.
\newblock pages 855--864, 2015.

\bibitem{o2008staging}
S.~O`Bryant et~al.
\newblock Staging dementia using clinical dementia rating scale sum of boxes
  scores: a [{Texas Alzheimer's} research consortium study.
\newblock {\em Archives of neurology}, 2008.

\bibitem{adni}
R.~Petersen et~al.
\newblock {Alzheimer`s Disease Neuroimaging Initiative (ADNI):} clinical
  characterization.
\newblock {\em Neurology}, pages 201--209, 2010.

\bibitem{ranganath2014black}
R.~Ranganath et~al.
\newblock Black box variational inference.
\newblock 2014.

\bibitem{sabuncu2011}
M.~Sabuncu et~al.
\newblock The dynamics of cortical and hippocampal atrophy in alzheimer
  disease.
\newblock pages 1040--1048, 2011.

\bibitem{stern2012cognitive}
Y.~Stern.
\newblock Cognitive reserve in ageing and {Alzheimer's} disease.
\newblock {\em The Lancet Neurology}, 11(11):1006--1012, 2012.

\bibitem{stock2003}
R.~Stock et~al.
\newblock The sigmoidal curve of cancer.
\newblock 2003.

\bibitem{dg2015}
H.~Suk et~al.
\newblock Hierarchical feature representation and multimodal fusion with deep
  learning for ad/mci diagnosis.
\newblock {\em Neuroimage}, 1(569-582), 2015.

\bibitem{tran2016edward}
D.~Tran et~al.
\newblock {Edward:} a library for probabilistic modeling, inference, and
  criticism.
\newblock {\em arXiv preprint arXiv:1610.09787}, 2016.

\bibitem{Xiang2013}
S.~Xiang et~al.
\newblock Multi-source learning with block-wise missing data for alzheimer's
  disease prediction.
\newblock pages 185--193, 2013.

\bibitem{daoqiangzhang2011}
D.~Zhang et~al.
\newblock Multimodal classification of alzheimer’s disease and mild cognitive
  impairment.
\newblock {\em NeuroImage}, 55(3):856--867, 2011.

\bibitem{ADkdd11}
J.~Zhou et~al.
\newblock Modeling disease progression via fused sparse group lasso.
\newblock {\em {SIGKDD}}, 2012.

\end{thebibliography}

\end{document}